\documentclass[10pt,twocolumn,letterpaper]{article}

\usepackage[pagenumbers]{cvpr} % To force page numbers, e.g. for an arXiv version

\definecolor{cvprblue}{rgb}{0.21,0.49,0.74}
\usepackage[pagebackref,breaklinks,colorlinks,allcolors=cvprblue]{hyperref}

\def\paperID{*****} % *** Enter the Paper ID here
\def\confName{CVPR}
\def\confYear{2025}

\title{Technical Report for the 5th CLVision Challenge at CVPR: Addressing the Class-Incremental with Repetition using Unlabeled Data - 4th Place Solution}

\author{\textbf{Panagiota Moraiti}$^{1*}$ \hspace{1.0em} \textbf{Efstathios Karypidis}$^{2,3*}$ \\
$^1$Tech Hive Labs
$^2$National Technical University of Athens   \\
$^3$Archimedes/Athena Research Center \\
\tt\small{panagiotamoraiti2@gmail.com, stathiskaripidis@gmail.com} \\
}

\begin{document}
\maketitle
\begin{abstract}
This paper outlines our approach to the 5th CLVision challenge at CVPR, which addresses the Class-Incremental with Repetition (CIR) scenario. In contrast to traditional class incremental learning, this novel setting introduces unique challenges and research opportunities, particularly through the integration of unlabeled data into the training process. In the CIR scenario, encountered classes may reappear in later learning experiences, and each experience may involve only a subset of the overall class distribution. Additionally, the unlabeled data provided during training may include instances of unseen classes, or irrelevant classes which should be ignored. Our approach focuses on retaining previously learned knowledge by utilizing knowledge distillation and pseudo-labeling techniques. The key characteristic of our method is the exploitation of unlabeled data during training, in order to maintain optimal performance on instances of previously encountered categories and reduce the detrimental effects of catastrophic forgetting. Our method achieves an average accuracy of 16.68\% during the pre-selection phase and 21.19\% during the final evaluation phase, outperforming the baseline accuracy of 9.39\%. We provide the implementation code at \href{https://github.com/panagiotamoraiti/continual-learning-challenge-2024}{https://github.com/panagiotamoraiti/continual-learning-challenge-2024}.
\end{abstract}
\vspace{-15pt}    
\section{Introduction}
The CLVision challenges \cite{clvision_workshop} are designed to promote research in visual continual learning \cite{wang2024comprehensive}, an emerging field of deep learning that focuses on enabling models to learn continuously from a stream of visual data. Participants are encouraged to address critical issues such as catastrophic forgetting \cite{Catastrophic-Forgetting} and adapting models to new tasks without requiring retraining from scratch.

Conventional deep learning models tend to suffer from catastrophic forgetting \cite{Catastrophic-Forgetting}, a problem where training on new classes causes them to lose the ability to recall earlier ones. Continual Learning, which is also known as Lifelong Learning or Incremental Learning, involves training a model incrementally on a stream of experiences in a continuous manner \cite{van2022three}. Class incremental learning \cite{CI} focuses on enabling a pre-trained model to learn new classes incrementally, which means that the model can extend its knowledge by integrating new categories over time without forgetting the previously learned ones.
In Class Incremental Learning with Repetition (CIR) \cite{CIR}, previously learned classes can reappear in later experiences, simulating a more realistic learning scenario. Each new experience may contain a mix of old and entirely new, unseen categories, representing only a subset of the overall class distribution.

This year's challenge  \cite{clvision_workshop} required participants to leverage unlabeled data during the training process. The unlabeled data may include instances of unseen or irrelevant classes, which should be ignored. To address this problem, we extend the baseline strategy, which already employs the well-known Learning without Forgetting (LwF) \cite{LWF, LWF_tpami} method for unlabeled data streams. Our approach also utilizes LwF for labeled data and incorporates the Less Forgetting Learning (LFL) \cite{LFL} strategy for both labeled and unlabeled data streams, aiming to retain previously learned knowledge and reduce catastrophic forgetting.
The exploitation of unlabeled data through a pseudo-labeling technique enables our method to maintain optimal performance on previously encountered categories, which the model tends to forget when trained on new experiences.

Our approach achieves an average accuracy of 16.68\% during the pre-selection phase and 21.19\% during the final evaluation phase, outperforming the baseline accuracy of 9.39\%. As a result, our method ranks 4th in the final evaluation.
\label{sec:intro}

\section{Methodology}
\label{sec:methodology}
\begin{figure*}[ht]
\centering
\includegraphics[width=\textwidth]{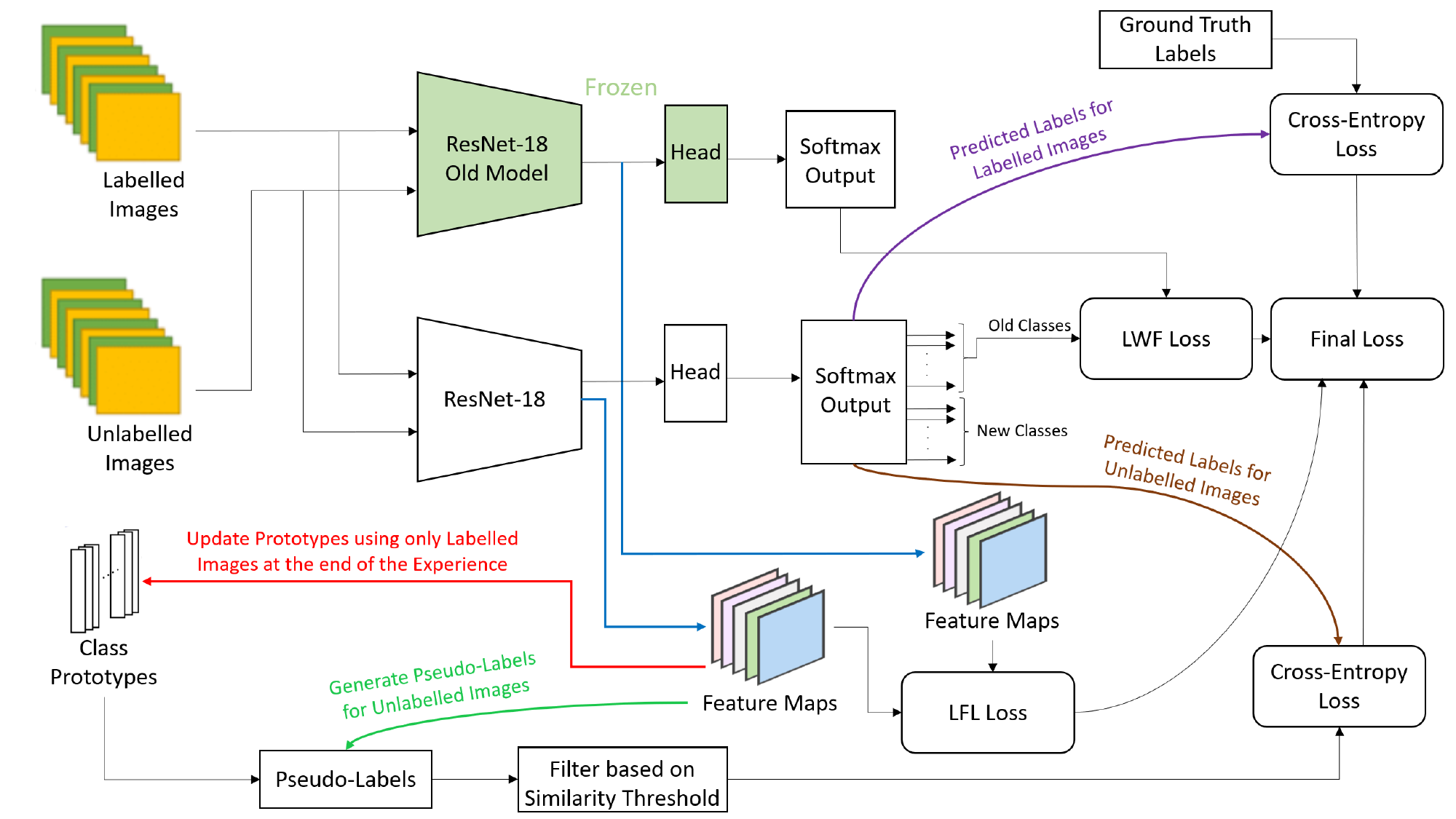}
\vspace{-5pt}
\caption{\textbf{An overview of our proposed method:} Each input image is passed through the model and a frozen instance of the model from the previous experience (old model). In addition to the standard \textbf{cross-entropy} loss between the model's predictions and ground truth labels, two additional losses are introduced to mitigate catastrophic forgetting. The \textbf{LwF} loss penalizes the difference in the logit outputs between the current model and the old model, while the \textbf{LFL} loss encourages feature similarity between the intermediate features of both models. As a final component, a \textbf{cross-entropy} loss is computed using the pseudo-labels generated for the unlabeled data, which are based on saved class prototypes. These representations are updated at the beginning of each experience, where the new ones are computed as the average of the old class prototypes and those from the most recent experience. All losses are combined to form the total loss. The overall method enables the model to effectively leverage both labeled and unlabeled data.}
\label{fig:overview}
\end{figure*}

Our approach consists of three main components: two focus on distilling knowledge \cite{hinton2015distilling} from models trained on previous experiences to the new one, while the third leverages unlabeled data for training through a pseudo-labeling technique \cite{yang2022survey}. An overview of our proposed method is provided in ~\cref{fig:overview}.

In our continual learning framework, we process both labeled samples $x_l \in \mathcal{X}_l$ and unlabeled samples $x_u \in \mathcal{X}_u$ at each experience $t$. The model parameters at experience $t$ are denoted as $\theta_t$, with $f_{\theta_{t}}$ representing the feature extractor and $g_{\theta_t}$ denoting the classification head. For the labeled data stream, following the baseline method, we employ standard supervised learning by minimizing the cross-entropy loss between the model's predictions and the ground truth labels. Let $y_l$ be the ground truth label for a labeled sample $x_l$. The classification loss is formulated as:
\begin{equation}
    \mathcal{L}_{sup} = -\sum_{i=1}^{C_t} \mathbb{1}[y_l = i] \log(p_i)
\end{equation}
where $p_i = \text{softmax}(g_{\theta_t}(f_{\theta_t}(x_l)))_i$ represents the predicted probability for class $i$, $C_t$ is the total number of classes seen up to experience $t$, and $\mathbb{1}[\cdot]$ is the indicator function that equals 1 when the condition is true and 0 otherwise.

Next we outline the principal modules of our proposed method. A crucial component involves maintaining a copy of the model, referred to as old model, that is exclusively trained on previous experiences, while the current model is being trained on new experience's data. At the end of each experience, the parameters of the retained model are updated to align with those of the current model.

\subsection{Maintaining Network's Capabilities}
The Learning without Forgetting (LwF) \cite{LWF,LWF_tpami} approach focuses on retaining knowledge from previous tasks while learning a new one. Unlike other replay-based methods \cite{bagus2021investigation}, which store examples from earlier tasks, LwF trains the network exclusively with new data that correspond to old categories, aiming at preserving its performance on prior tasks. It functions as a regularizer by minimizing the discrepancy between the model's output on old tasks before and after learning the new task. 

The model in every new experience includes extra nodes in the output layer to accommodate the new classes, which are fully connected to the previous layer with weights that are randomly initialized. Experiments have shown that if the outputs of the old model on new data are similar to the outputs of the new model, then the capabilities on previous tasks are maintained \cite{LWF,LWF_tpami}. This means that the model is able to integrate new knowledge without forgetting previously learned classes.

In our approach, we extend the baseline strategy, which already employs the LwF method for unlabeled data streams, by also incorporating labeled data into the LwF framework. Since LwF utilizes Knowledge Distillation, it does not require labels and therefore all available data can be utilized to mitigate the negative effects of catastrophic forgetting. The Knowledge Distillation term is defined as the the Kullback-Leibler (KL) divergence between the current and old model's logits:
\begin{equation}
    \text{L}_{KD}(\mathbf{z}, \mathbf{z}^{\text{old}}) = \text{KL}\left(\text{S}\left(\frac{\mathbf{z}}{T}\right) \parallel \text{S}\left(\frac{\mathbf{z}^{\text{old}}}{T}\right)\right)
\end{equation}
where \( z \) and \( z_{old} \) are the logit outputs of the current and old model, \( S \) is the softmax function and \( T \) is the temperature parameter.

The total LwF loss, denoted as \( L_{\text{LwF}} \) combines the contributions from both labeled and unlabeled data streams:
\begin{equation}
    L_{\text{LwF}} = \alpha_l \cdot \text{L}_{KD}(\mathbf{z}_l, \mathbf{z}_l^{\text{old}}) + \alpha_u \cdot \text{L}_{KD}(\mathbf{z}_u, \mathbf{z}_u^{\text{old}})
\end{equation}
where \(\alpha_l\) and \(\alpha_u\) are the weights for the distillation losses applied to labeled and unlabeled data, respectively. The notations $z_l =(g_{\theta_t}(f_{\theta_t}(x_l)))$ and $z_l^{old}$ refer to the logit outputs of the current and previous model for labeled data, while $z_u =(g_{\theta_t}(f_{\theta_t}(x_u)))$ and $z_u^{old}$ denote the corresponding outputs for unlabeled data.

\subsection{Preserving Feature Consistency}
The Less Forgetting Learning (LFL) \cite{LFL} strategy is similar to LwF, but instead of preventing changes in the outputs of the new model, it focuses on minimizing changes in the model's intermediate representations during the learning process.

Inspired by LFL, our method leverages the representations of both labeled and unlabeled data. In each new experience, features are extracted from the new and old models, and discrepancies between them are penalized. Unlike the approach proposed in LFL, none of the network's weights are frozen. We observed that while freezing certain layers helps prevent catastrophic forgetting, it also results in a significant degradation of the model's performance on new classes. We implement this strategy using the Mean Squared Error (MSE) loss, applied to both labeled and unlabeled data streams. The LFL loss, denoted as \( L_{\text{LFL}} \) is formulated as:
\begin{equation}
    L_{\text{LFL}} = \beta \cdot (\text{MSE}(\mathbf{h}_l, \mathbf{h}_l^{\text{old}}) +  \text{MSE}(\mathbf{h}_u, \mathbf{h}_u^{\text{old}}))
\end{equation}
where \(\beta\) is the weight for the Mean Squared Error losses on both the labeled and unlabeled data streams. The terms \( \mathbf{h}_l\) and \( \mathbf{h}_l^{\text{old}}\) indicate the feature representations of the labeled data from the current and previous models, while \( \mathbf{h}_u \) and \( \mathbf{h}_u^{\text{old}} \) refer to the correspoding representations of the unlabaled data.

\subsection{Prototype Generation and Pseudo-Labeling Technique}
Our method involves generating prototypes from the labeled data stream at the beginning of each experience to facilitate pseudo-labeling of unlabeled data. For each class \( c \), a prototype \( \mathbf{p}_c \) is computed as the average of all feature vectors \( \mathbf{h}_l \) associated with that class:
\begin{equation}
    \mathbf{p}_c = \frac{1}{N_c} \sum_{i=1}^{N_c} \mathbf{h}_{l, i}
\end{equation}
where \( N_c \) is the number of labeled samples in class \( c \). These prototypes, along with their corresponding labels, are stored in a buffer for direct reference. At the beginning of each experience they are updated as the average of the old prototypes and those from the most recent experience.

For each unlabeled sample, a pseudo-label \(\hat{y}_u\) is assigned, based on the highest cosine similarity between its features \( \mathbf{h}_u \) and each class prototype:
\begin{equation}
    \hat{y}_u = \arg\max_c \frac{\mathbf{h}_u \cdot \mathbf{p}_c}{\|\mathbf{h}_u\| \|\mathbf{p}_c\|}
\end{equation}
A pseudo-label is only assigned if this similarity exceeds a predefined threshold \(\tau\), ensuring that the sample doesn't belong to an unseen or distractor class not present in the labeled data.

When pseudo-labels are assigned, a cross-entropy loss is applied:
\begin{equation}
    \mathcal{L}_{\text{pseudo}} = - \gamma \cdot \sum_{i=1}^{U} \mathbb{1}[\hat{y}_u = i] \log(p_i)
\end{equation}
where \( p_i =  \text{softmax}(g_{\theta_t}(f_{\theta_t}(x_u)))_i \) is the predicted probability for pseudo-labeled class \( i \), \(\gamma\) is the weight for pseudo-labeled loss and \( U \) is the number of unlabeled samples. This approach enables us to leverage previously encountered classes from the labeled data stream when they appear in the unlabeled data in future experiences, thereby expanding the dataset available for training. 

The total loss function is a combination of the classification, distillation, Less Forgetting Learning (LFL), and pseudo-labeling losses. It is expressed as:
\begin{equation}
    \mathcal{L}_{\text{total}} = \mathcal{L}_{\text{sup}} + L_{\text{LwF}} + L_{\text{LFL}} + \mathcal{L}_{\text{pseudo}}
\end{equation}
This comprehensive loss function ensures that the model effectively learns from both labeled and unlabeled data, while maintaining past knowledge and expanding its capability to incorporate new classes.
\section{Experiments}
\label{sec:experiments}
\subsection{Experimental Setup}
\paragraph{Data.} The experiments were conducted on an ImageNet-like computer vision dataset comprising 130 classes, structured in a Class-Incremental with Repetition (CIR) learning framework. The dataset presents 50 sequential experiences, each containing both labeled (500 images) and unlabeled (1,000 images) data streams, with balanced classes within each experience. Of the total categories, 100 are designated as learnable, while 30 serve as distractors. The challenge explores three scenarios of increasing complexity: Scenario 1, where labeled and unlabeled streams contain identical class distributions; Scenario 2, where the unlabeled stream includes samples from all current and future learnable classes; and Scenario 3, which adds samples from the 30 distractor categories to the unlabeled stream while maintaining the structure from Scenario 2. 

Evaluation is performed on a balanced test set containing novel instances from all previously seen classes, excluding distractor classes. This incremental setup allows for comprehensive assessment of continual learning strategies under varying degrees of unlabeled data complexity.
\paragraph{Implementation Details.} 
Our implementation is built on top of challenge DevKit which is based on Avalanche library \cite{avalanche,avalanche_cvpr}. We utilize the ResNet-18 \cite{he2016deep} as the backbone network, as mandated by the competition guidelines, and extends the baseline Learning without Forgetting (LwF) approach by applying it to both labeled and unlabeled data streams. We employ Adam \cite{adamopt} optimizer with a learning rate of 5e-4 and implement a StepLR scheduler with a step size of 5 and the gamma equals to 0.5 for learning rate decay. The training batch size is set to 32, while the test batch size is 256. 

To leverage unlabeled data effectively, we maintain a buffer of size 100 to store class prototypes, each with a feature dimension of 512, along with their corresponding labels. For pseudo-labeling of unlabeled samples, we employ a confidence threshold of $\tau=0.5$ and samples below this threshold are excluded to prevent contamination from unseen or distractor classes. Regarding loss term weights we use $a_l=2, a_u=2, \beta=1000 $ and $\gamma=0.25$. To prevent overfitting to current classes, we limit the training to a maximum of 15 epochs per experience and implement early stopping based on validation performance.

\subsection{Results}
In ~\cref{tab:results}, we present the results from our study, showcasing the improvements achieved by our approach across the three distinct scenarios and two phases: Pre-selection and Final evaluation. Our results are compared with a baseline method provided by the organizers of the challenge. In the pre-selection phase, we observe a significant improvement in average accuracy, which increases by +7.29\%, reaching 16.68\%. In the final evaluation phase, our approach shows further improvement, with the average accuracy rising by +11.80\%, reaching 21.19\%. As a result, our method ranks 4th in the final evaluation.

Notably, we observe significant performance enhancements in Scenarios 2 and 3. This can be explained by the fact that the unlabeled data in these scenarios may include samples from classes that have been encountered before, which helps the model to remember them and mitigate forgetting. By using Learning without Forgetting (LWF) and Less Forgetting Learning (LFL) on both labeled and unlabeled data streams, our approach effectively minimizes changes in the model's outputs and intermediate feature representations, thereby reducing the issue of forgetting. Moreover, by assigning pseudo-labels to unlabeled data based on their similarity to class prototypes, we can leverage prior knowledge to sustain high performance on old tasks.  Our results highlight the effectiveness of integrating these techniques to enhance the learning process, demonstrating the value of utilizing unlabeled data to reinforce previously acquired knowledge.

\renewcommand{\arraystretch}{1.15}
\begin{table}[t!]
\normalsize
\centering
\setlength{\tabcolsep}{3.0pt}
\resizebox{\columnwidth}{!}{%
\begin{tabular}{lcccc}
\toprule
& \multicolumn{4}{c}{\textbf{Results}} \\
\cmidrule(r){2-5}
\textbf{Phase} & \textbf{Scenario 1} & \textbf{Scenario 2} & \textbf{Scenario 3} & \textbf{Average Accuracy} \\
\midrule
\textbf{Baseline} & 7.96 & 10.66 & 9.54 & 9.39 \\
\hline
\hline
\textbf{Pre-selection} & 14.42 (+6.46\%) & 19.02 (+8.36\%) & 16.60 (+7.06\%) & \textbf{16.68 (+7.29\%)} \\
\textbf{Final} & 17.49 (+9.53\%) & 22.44 (+11.78\%) & 23.64 (+14.10\%) & \textbf{21.19 (+11.80\%)} \\
\bottomrule
\end{tabular}%
}
\vspace{-5pt}
\caption{\textbf{Performance across different scenarios and phases.}}
\label{tab:results}
\vspace{-7pt}
\end{table}

\section{Conclusion}
\label{sec:conclusion}
In this paper, we presented our approach to the 5th CLVision Challenge at CVPR, focused on addressing the Class-Incremental with Repetition (CIR) scenario using unlabeled data. Our method successfully tackles catastrophic forgetting by combining three key components: applying Learning without Forgetting (LwF) to both labeled and unlabeled data streams, implementing Less Forgetting Learning (LFL) to preserve feature consistency across model iterations, and leveraging a prototype-based pseudo-labeling technique to effectively utilize unlabeled data. The experimental results demonstrate the effectiveness of our approach. Our method shows particular strength in Scenarios 2 and 3, where the unlabeled data contains samples from previous or future learnable classes and distractor classes, highlighting the value of our pseudo-labeling strategy in complex learning environments. This work contributes to the field of continual learning by showcasing how unlabeled data can be effectively leveraged to maintain performance on previously encountered categories while accommodating new knowledge. 

Future work could explore more sophisticated prototype generation techniques, adaptive threshold mechanisms for pseudo-labeling, and integration with memory-efficient replay methods to further enhance performance in challenging continual learning scenarios.
{
    \small
    \bibliographystyle{ieeenat_fullname}
    \bibliography{main}
}

% WARNING: do not forget to delete the supplementary pages from your submission 
% \input{sec/X_suppl}

\end{document}